\documentclass[10pt,twocolumn,letterpaper]{article}

\usepackage{cvpr}              % To produce the CAMERA-READY version
\usepackage{booktabs}
\usepackage{multirow}

\author{
Zhenghui Guo$^{1}$ \quad
Yuanbin Man$^{2}$ \quad
Junyuan Sheng$^{5}$ \quad
Bowen Lin$^{1}$ \quad
Ahmed Ahmed$^{1}$ \quad
Bo Jiang$^{4}$ \\
Boyuan Zhang$^{3}$ \quad
Miao Yin$^{2}$ \quad
Sian Jin$^{4}$ \quad
Omprakash Gnawali$^{1}$ \quad
Chengming Zhang$^{1}$\thanks{Corresponding author.}
\\[0.8em]
$^{1}$University of Houston \quad
$^{2}$The University of Texas at Arlington \quad
$^{3}$Indiana University Bloomington \\
$^{4}$Temple University\quad
$^{5}$Independent Researcher
}

\usepackage{algorithm}
\usepackage{algorithmicx}
\usepackage{algpseudocode}

\usepackage{tabularx}

\DeclareUnicodeCharacter{2265}{\ensuremath{\geq}} % ≥
\DeclareUnicodeCharacter{202F}{\,}               % 
\usepackage{pifont}
\newcommand{\cmark}{\ding{51}} % ✓
\newcommand{\xmark}{\ding{55}} % ✗
\usepackage{placeins}
\usepackage{float}

\usepackage{caption}
\usepackage{graphicx}
\usepackage{subcaption}
\usepackage[table]{xcolor}

\usepackage{algorithm}
\usepackage{algpseudocode}
\usepackage{amsmath, amssymb}

\algrenewcommand\algorithmicrequire{\textbf{Input:}}
\algrenewcommand\algorithmicensure{\textbf{Output:}}

\usepackage[normalem]{ulem} 
\usepackage{xcolor}

\usepackage{microtype}

\usepackage{tikz}
\newcommand*\Circled[2][gray!40]{% require `tikz`
	\tikz[baseline=(char.base)]{\node[
        shape=circle, draw=none,  thick, 
        fill=#1 ,inner sep=0.9pt] (char) 
    {\textcolor{black}{#2}}; 
}}

\renewcommand{\paragraph}[1]{\vspace{.5em}\noindent\textbf{#1.}}

\usepackage{xspace}

\usepackage{soul}
\setuldepth{foobar}

\definecolor{cvprblue}{rgb}{0.21,0.49,0.74}
\usepackage[pagebackref,breaklinks,colorlinks,allcolors=cvprblue]{hyperref}

\def\paperID{*****} % *** Enter the Paper ID here
\def\confName{CVPR}
\def\confYear{2026}

\title{Event-VStream: Event-Driven Real-Time Understanding for Long Video Streams}

\begin{document}
\maketitle
\begin{abstract}
Real-time understanding of long video streams remains challenging for multimodal large language models (VLMs) due to redundant frame processing and rapid forgetting of past context. Existing streaming systems rely on fixed-interval decoding or cache pruning, which either produce repetitive outputs or discard crucial temporal information. We introduce Event-VStream, an event-aware framework that represents continuous video as a sequence of discrete, semantically coherent events. Our system detects meaningful state transitions by integrating motion, semantic, and predictive cues, and triggers language generation only at those boundaries. Each event embedding is consolidated into a persistent memory bank, enabling long-horizon reasoning while maintaining low latency. Across OVOBench-Realtime, and long-form Ego4D evaluations, Event-VStream achieves competitive performance. It improves over a VideoLLM-Online-8B baseline by +10.4 points on OVOBench-Realtime, achieves performance close to Flash-VStream-7B despite using only a general-purpose LLaMA-3-8B text backbone, and maintains around 70\% GPT-5 win rate on 2-hour Ego4D streams.

\end{abstract}
   
\section{Introduction}
\label{sec:intro}

Understanding and responding to long-form real-time video streams is essential for the development of next-generation AI systems. Applications, e.g., AR/VR assistants, home robots, autonomous vehicles, and content moderation systems, require multimodal models~\cite{openai2023gpt4v, gemini15, wang2024qwen2vl} that can perceive, retain, and react efficiently. These systems must operate in dynamic environments, maintain temporal context, and deliver responses with low latency.

\begin{figure}[th]
    \centering
    \includegraphics[width=1.0\linewidth]{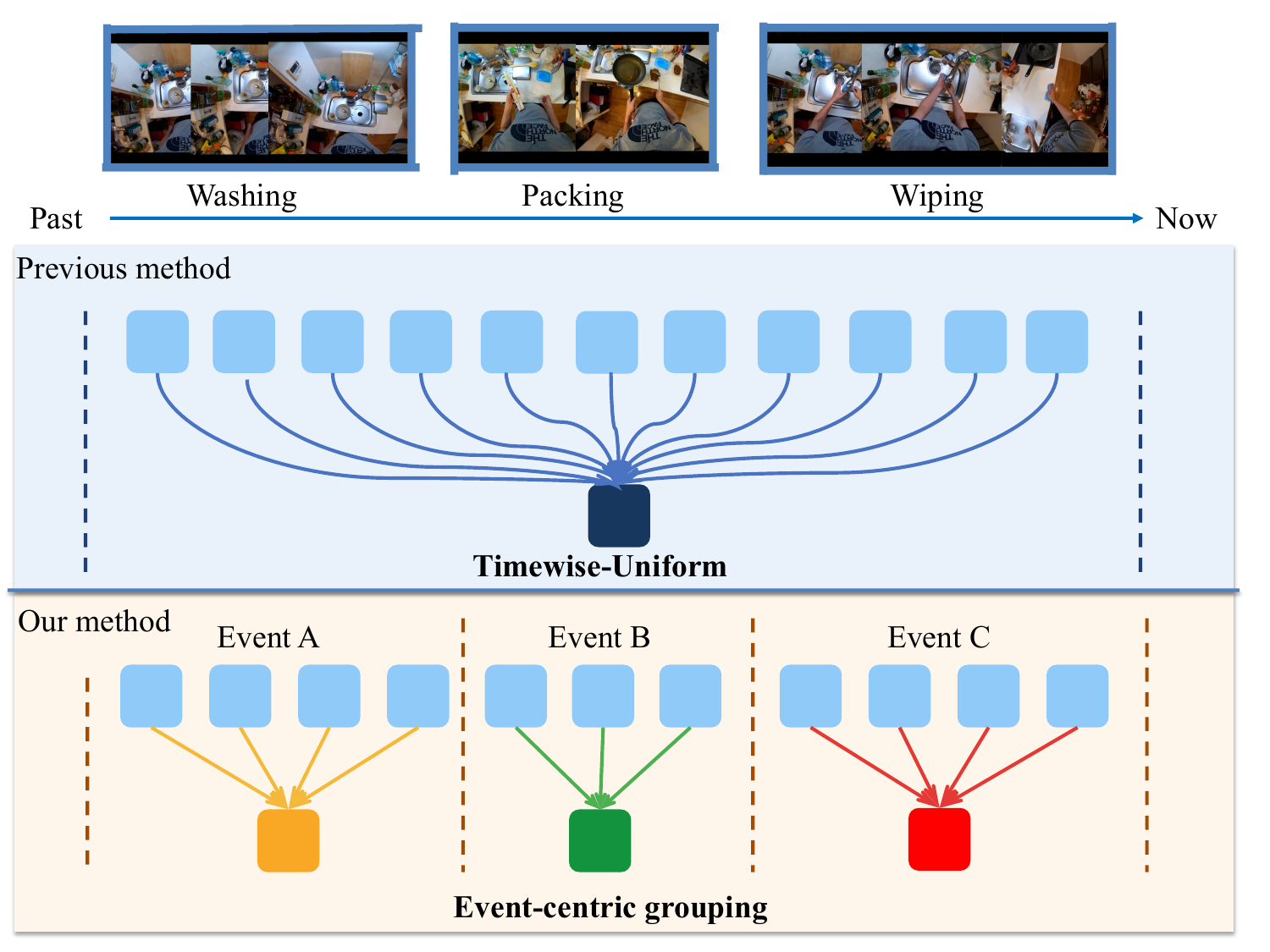}
    \caption{Comparison between timewise-uniform processing and our event-centric grouping. Previous streaming models treat every frame equally over time, leading to redundant computation and temporally fragmented context. In contrast, our method dynamically clusters frames into semantically coherent events (A–C), processing and updating memory only when meaningful visual changes occur.}
    \vspace{-6mm}
    \label{fig:placeholder}
\end{figure}

Unlike offline video understanding, where models have access to the entire video sequence, streaming video-language models (VLMs)~\cite{chen2024videollmonline, xu2025streamingvlm, zhang2024flashvstream} must process unbounded video streams in an online manner, without access to future frames, while effectively retaining memory of past events. In this context, an event denotes a temporally coherent segment that captures meaningful visual or semantic changes within the video.

\begin{figure*}[t!]
    \centering
    \includegraphics[width=\textwidth, trim=10 5 10 5, clip]{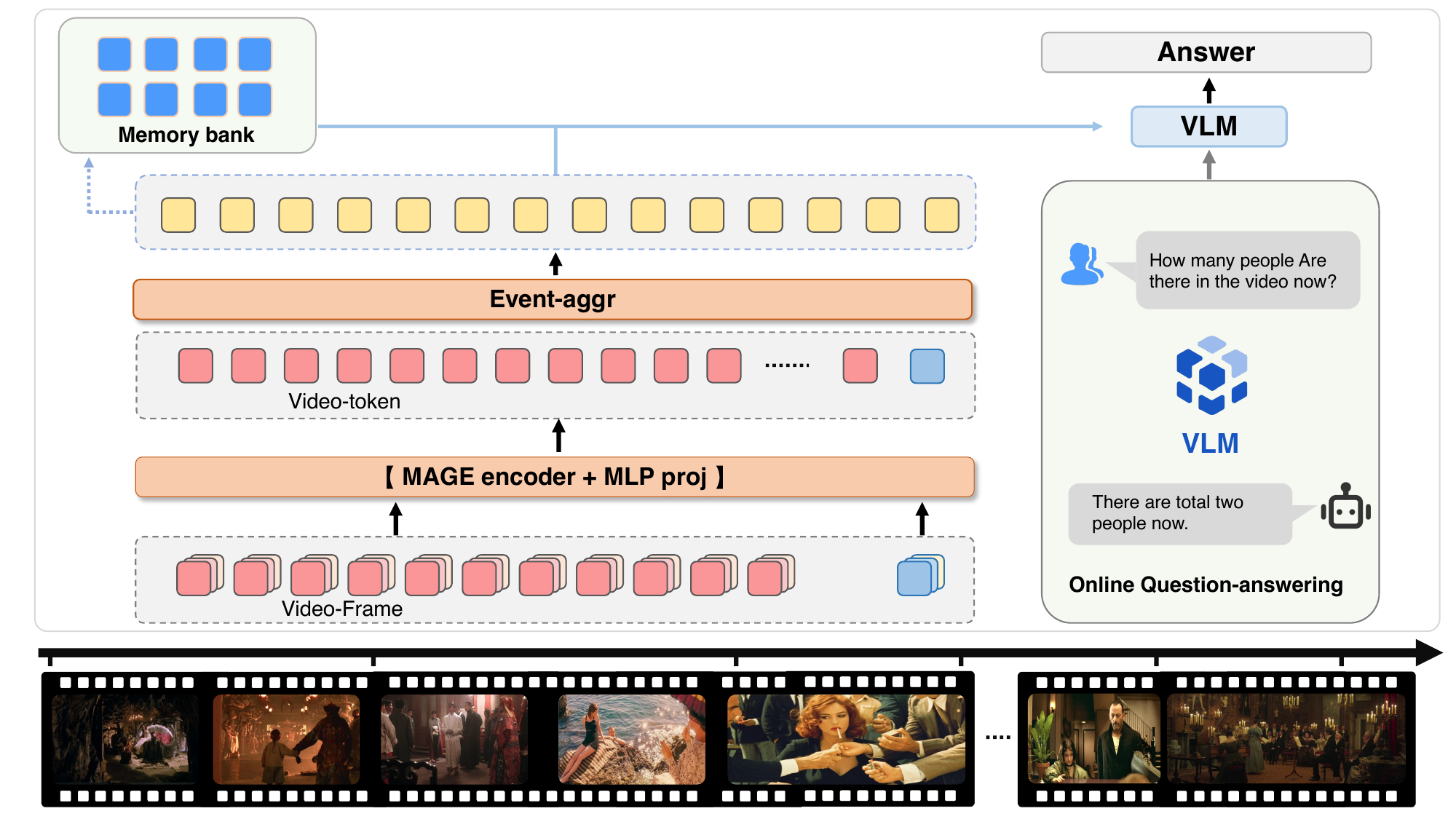}
    \vspace{-2mm}
    \caption{
        \textbf{Overview of the proposed Event-VStream.}
        Our system dynamically groups continuous video frames into semantically coherent \textit{events}.
        Each event embedding is compressed and stored in a persistent \textit{memory bank}\cite{zhang2024flashvstream,xiong2025streamchat},
        enabling efficient long-horizon reasoning and online question answering under streaming conditions.
        The model integrates motion-based and semantic cues for event aggregation, retrieves relevant event memories, and performs event-driven decoding to maintain temporal coherence.}
    \vspace{-3mm}
    \label{fig:framework}
\end{figure*}

However, existing approaches are hindered by two key challenges: \textit{redundancy} and \textit{forgetting}. Generally, high-frequency decoding generates responses at every frame to ensure real-time output, yet often produces nearly identical predictions, leading to redundant computation and limited informativeness~\cite{xu2025streamingvlm}. To address the infinite frames, existing systems typically refresh or sparsify the KV cache at fixed intervals~\cite{di2025rekv,wan2024look,wan2025meda,yang2025streammem}. This strategy mitigates memory growth at the language-model level only after redundant visual tokens have already been processed. Still, it disrupts temporal continuity and discards past information before it can be semantically consolidated~\cite{wang2025retake}. As a result, current strategies reduce memory usage but fail to prevent redundant visual embeddings from entering the model in the first place, making it difficult to jointly minimize redundancy and prevent forgetting of relevant information in real-time video understanding.

\vspace{0.5em}

These limitations indicate that the challenges lie not only in memory retention, but also in the representation of the video stream itself before it is encoded into the LLM~\cite{zhang2024flashvstream,jiang2025storm,wang2025metok,kim2025infinipotv}, which can lead to massive redundancy.
Beyond these KV-cache-based strategies~\cite{ning2025livevlm,xu2025streamingvlm,di2025rekv},
most frame-level approaches~\cite{zhang2024flashvstream,tang2025adaptive,Zhang2024LongContextTransfer}
still process videos as uniformly sampled or compressed frame embeddings,
implicitly assuming that visual dynamics evolve smoothly over time
and that fixed-size temporal windows are sufficient to capture meaningful context. However, this assumption does not align with the non-uniform, event-driven dynamics of real-world videos and human perception~\cite{shou2021gebd,Mounir2023STREAMER}.As illustrated in Figure~~\ref{fig:placeholder}, timewise-uniform processing treats every frame equally and leads to redundant computation and fragmented temporal context, whereas our event-centric grouping updates memory only when meaningful state changes occur.

Empirically, video semantics tend to remain stable over intervals and then change abruptly, forming natural event boundaries\cite{shou2021gebd}. Insights from cognitive science further show that humans segment continuous experience into discrete events~\cite{zacks2010event,kurby2008segmentation}, and update mental models primarily when predictions fail\cite{Mounir2023STREAMER}, rather than at uniform time steps. This reveals a fundamental misalignment: time-uniform windows impose artificial boundaries that disrupt semantic coherence and fail to reflect how meaning actually unfolds. This leads to our core question: \textit{How should VLMs represent and reason over streaming video in a way that avoids redundancy, preserves memory, and aligns with how humans perceive the world?}

This leads to a key insight: humans perceive and understand the world not as a continuous stream of frames or fixed slices, but as a sequence of discrete events. Representing video in terms of events, rather than frames, better aligns with how both perception and semantics evolve over time. Inspired by this, we propose an event-centric perspective for streaming video understanding, where video is modeled as a dynamic sequence of meaningful state transitions rather than a uniform signal. This paradigm enables models to update only when the world changes, maintain abstract memory over events, and generate contextually rich outputs in real time.

In summary, the proposed event-centric strategy offers a more human-aligned representation of streaming video, enabling selective updates, long-term semantic memory, and coherent real-time reasoning. To operationalize this perspective, we introduce \textbf{Event-VStream}, a framework equipped with three core capabilities. We summarize our contributions as follows: 

\begin{itemize}
    \item We develop an \textbf{event boundary detector} that integrates motion, semantic drift, and prediction cues to convert continuous frames into compact, boundary-aware event representations.
    \item We introduce a lightweight \textbf{event-level memory bank} that merges redundant events and provides persistent, non-redundant context for long-horizon streaming reasoning.
    \item We propose an \textbf{event-triggered decoding strategy} that generates text only at detected semantic transitions with simple pacing control, which reduces redundant updates and, together with our event-centric representation and memory, maintains coherent real-time narration over multi-hour streams.
    \item  Extensive experiments show that Event-VStream maintains over 70\% GPT-5 win rate on 2-hour Ego4D streams, improves OVOBench-Realtime performance by +10.4 points over its VideoLLM-Online backbone, and sustains sub-0.1\,s/token real-time latency.
\end{itemize}

\section{Related Work}
\label{sec:related}

% All text must be in two-column format.
% The total allowable size of the text area is $6\frac78$ inches (17.46 cm) wide by $8\frac78$ inches (22.54 cm) high.
% The columns should be $3\frac14$ inches (8.25 cm) wide, with a $\frac{5}{16}$ inch (0.8 cm) space between them.
% The main title (on the first page) should begin 1 inch (2.54 cm) from the top edge of the page.
% The second and following pages should begin 1 inch (2.54 cm) from the top edge.
% On all pages, the bottom margin should be $1\frac{1}{8}$ inches (2.86 cm) from the bottom edge of the page for $8.5 \times 11$-inch paper;
% for A4 paper, approximately $1\frac{5}{8}$ inches (4.13 cm) from the bottom edge of the
% page.

\subsection{Vision-Language Model (VLMs)} 

Recent advances in vision language models (VLMs)\cite{alayrac2022flamingo,li2022blip,li2023blip2,liu2023llava,liu2024llavanext} have substantially improved multimodal reasoning across both static and dynamic scenes. Flamingo~\cite{alayrac2022flamingo} bridges pretrained vision-only and language-only models, allowing multimodal reasoning\cite{alayrac2022flamingo,li2022blip,li2023blip2} over interleaved image–text inputs. In contrast, the BLIP family~\cite{li2022blip,li2023blip2,dai2023instructblip} introduce a unified architecture that aligns cross-modal features through contrastive, matching, and generative objectives, leading to powerful vision-language model pretraining. Moreover, the LLaVA series~\cite{liu2023llava,liu2024llavanext} extends large language models with visual instruction tuning, which supports image-grounded dialogue and reasoning. However, although these methods achieve strong results on short-term video understanding, they struggle to maintain coherence and memory over long-term or streaming videos.

% While these models mark major progress in multimodal understanding, they operate on static visual inputs and do not address the challenges of temporal redundancy and long-range semantics that arise in continuous video streams.

% Vision Language Models (VLM) is a type of multimodal artificial intelligence model designed to jointly process and reason over visual and linguistic information, Flamingo \cite{alayrac2022flamingo} effectively bridge powerful pretrained vision-only and language-only models, handle sequences of arbitrarily interleaved visual and textual data, and seamlessly ingest images or videos as inputs, BLIP series  \cite{li2022blip,li2023blip2,dai2023instructblip} introduces a unified vision–language architecture combining contrastive, matching, and generative objectives within a multimodal encoder–decoder framework. It jointly aligns visual–text features (ITC), learns cross-modal interactions (ITM), and generates image-grounded text (LM) for comprehensive vision–language pretraining, The LLaVA series \cite{liu2023llava,liu2024llavanext} integrates a pretrained visual encoder with large language models through visual instruction tuning, enabling image-conditioned dialogue and reasoning, researchers extended image data to video. While these models introduced a breakthrough in VLM, it did not try to solve the  new challenges of redundant frame features in long videos. 

\subsection{Long Video Understanding}
Existing approaches enable long-term video understanding by reducing temporal redundancy. MIST~\cite{Gao_2023_CVPR} selects question-relevant video segments, while SEVILA~\cite{Yu2023SelfChained} jointly performs key-frame localization and video question answering (VideoQA), making the filtering process query-aware. These methods are effective for VideoQA; however, they rely on question supervision and do not generalize to open-ended streaming scenarios. MovieChat~\cite{Song_2024_CVPR_MovieChat} aggregates similar frames via average pooling to fit long videos into limited GPU memory, although its training-free design results in semantic loss. Long-context fine-tuning methods, e.g., V-NIAH~\cite{Zhang2024LongContextTransfer}, expand the context capacity of VLMs while remaining computationally expensive and restricted to frame-level representations. RETAKE~\cite{wang2025retake} prunes the KV cache to reduce redundancy, yet remains token-centric and neglects higher-level temporal structure. Besides token selection and compression, memory-augmented models such as MA-LMM further introduce explicit long-term memory modules for hour-level video understanding \cite{he2024ma}. These works address redundancy by filtering or compressing visual tokens within a time- or frame-based paradigm. In contrast, we adopt an event-centric representation that retains only meaningful state transitions as the fundamental unit for streaming understanding.

\subsection{Streaming Video-Language Models}
Multimodal LLMs\cite{bai2025qwen25vl,liu2023llava,alayrac2022flamingo} have been extended from offline video understanding to real-time streaming. VideoLLM-Online~\cite{chen2024videollmonline} introduces the streaming framework to enable temporally aligned, long-context video conversation. StreamingVLM~\cite{xu2025streamingvlm} further aligns training and streaming inference for stable real-time understanding of unbounded visual input. LiveVLM~\cite{ning2025livevlm} proposes a training-free streaming paradigm based on KV-cache pruning and frame-wise merging, and StreamChat~\cite{xiong2025streamchat} leverages hierarchical memory to support multi-round video dialogue. VideoStreaming~\cite{qian2024VideoStreaming}, an advanced vision-language large model (VLLM) for video understanding that can process arbitrary-length videos with a constant number of adaptively selected video tokens in a streaming manner. Despite enabling real-time inference, these approaches still suffer from inconsistent context understanding across long~\cite{cheng2024enhancing,wang2025episodic} or streaming videos and in overcoming computational bottlenecks for extended video processing.

% Despite these advances, existing systems remain time-centric and struggle in long-video scenarios: they either lose distant context, fail to answer temporally grounded queries, or accumulate increasingly complex memory structures that degrade efficiency over time.
% introduced multi-round, conversational video reasoning via a hierarchical memory system that merges short-term, long-term, and dialogue memories. However, these methods suffer from performance degradation in the long-video context; they either cannot sustain reasoning over long video, lose distant context, lose the ability to answer temporal and spatial questions for the old context, or grow hierarchal complex memory over time. 

\subsection{Efficient Video Token Decoding}
To mitigate visual redundancy, token-efficient long video understanding for multimodal LLMs~\cite{jiang2025storm} incorporates a temporal encoder between the vision encoder and the Large Language Model (LLM), filtering out uninformative patches and achieving up to 8$\times$ computational speedups. Flash-VStream~\cite{zhang2024flashvstream} further introduces Spatial-Temporal-Abstract-Retrieved (STAR) memory, a learnable module that compresses frame features via temporal-weighted clustering to maintain compact representations during streaming inference. However, these approaches primarily operate at the token or patch level and rely on static compression policies that do not adapt to scene dynamics or user query semantics, limiting their effectiveness in long video streaming scenarios.

% \textbf{Efficient KV Cache Compression and Merging }
% Long-context LLMs face GPU bottlenecks because KV cache size grows linearly with sequence length and it grows a lot faster for Multimodal LLMs , Model Tells You Where to Merge (KVMerger) \cite{wang2024kvmerger} proposes adaptive KV merging based on token-level cosine similarity, It identifies merging sets via Agglomerative Hierarchical Clustering Algorithm and applies Gaussian kernel weighted merging algorithm to selectively merge all states within each merging set, Efficient Long-Context LLM Inference via KV Cache Clustering (Chelsea) \cite{hu2025chelsea} introduces Chunked Soft Matching, a lightweight KV clustering method that groups tokens within chunks using alternating partition rules and merges them into centroids, However these methods add computational overhead duo to similarity calculations at inference time, it also assumes uniform local similarity and can mis-cluster tokens and does not adapt compression ration dynamically to sematic complexity or query load. 

\section{Why Events? Empirical and Cognitive Foundations}
% Through analysis across diverse long-form videos, we observe that video streams naturally unfold in events rather than in smooth temporal trajectories. We present three complementary findings that motivate an event-centric formulation

We address redundancy and forgetting in streaming video understanding by representing continuous video as a sequence of discrete events. Event boundaries are detected through motion and semantic cues, and each event embedding is stored in a persistent memory. Language decoding is triggered only when meaningful changes occur, enabling selective updates and coherent long-horizon reasoning over unbounded streams.

\subsection{Empirical Finding: Video Structure is Event-Centric Rather Than Frame-Sequential}

\begin{figure}[t]
    \centering
    \hspace*{-2mm}
    \begin{subfigure}[c]{0.455\linewidth}
        \centering
        \includegraphics[height=3.85cm, clip]{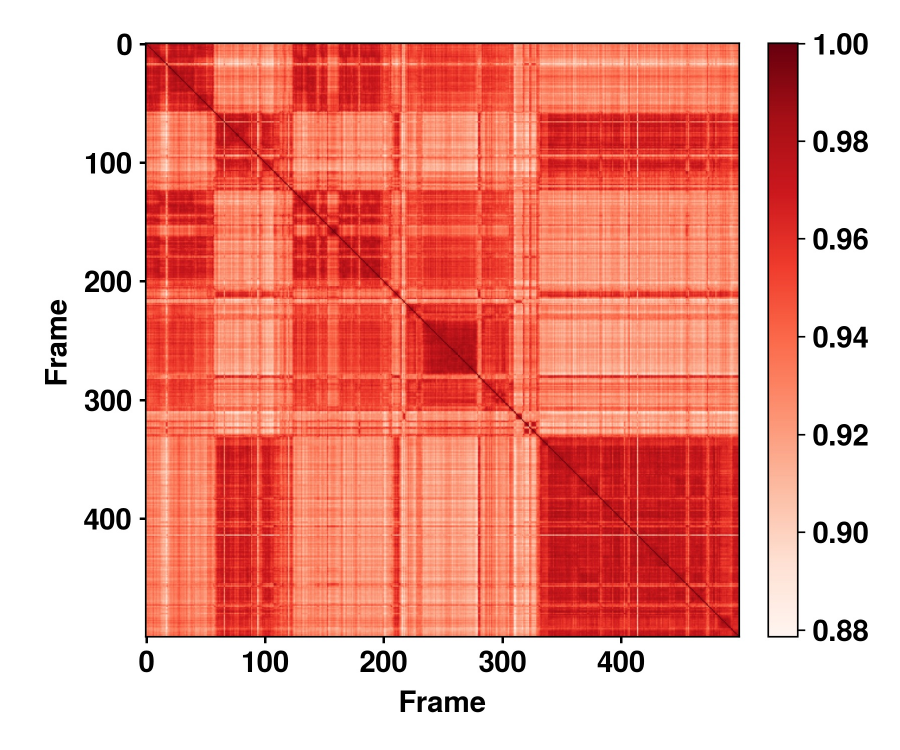}
        \caption{Frame-level embedding similarity matrix. }
        \label{fig:frame_similarity_matrix_a}
    \end{subfigure}
    \hfill
    % \hspace{0.012\linewidth}
    \begin{subfigure}[c]{0.455\linewidth}
        \centering
        \includegraphics[height=3.7cm, clip]{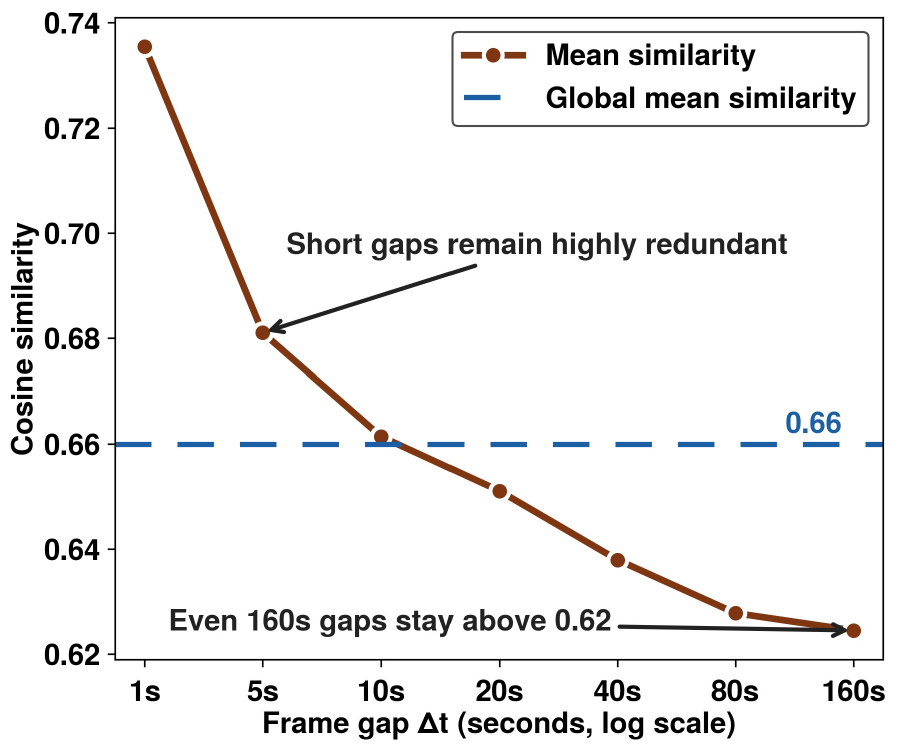}
        \caption{Mean cosine similarity decay over increasing frame gaps.}
        \label{fig:frame_similarity_matrix_b}
    \end{subfigure}
    % \vspace{2pt}
    \caption{
        (\ul{\textbf{a}}) Visual embeddings form block structures, revealing event-level recurrence rather than smooth temporal evolution. 
        (\ul{\textbf{b}}) Temporal redundancy decays nonlinearly with time, reinforcing the need for Event-VStream compression rather than naïve pooling.
    }
    \label{fig:frame_similarity_matrix_combined}
\end{figure}

Our analysis of frame-level embedding similarity reveals a block-structured recurrence (see Figure~\ref{fig:frame_similarity_matrix_a}), where semantically coherent segments reappear over time rather than evolving smoothly frame by frame. More importantly, temporal redundancy decreases sharply only at event boundaries (see Figure~\ref{fig:frame_similarity_matrix_b}), instead of decaying gradually over time. These findings indicate that fixed-length windows are misaligned with the true semantic structure of video, motivating the design of our proposed \textbf{Event-VStream}.

% %TODO ZHNEGHUI
% \subsection{Quantitative Finding: Event Segmentation Better Preserves Semantics}

% \TODO{Do we refer to any figure?} We compare three common streaming representations—overlapping windows, non-overlapping chunks, and our event-based segmentation—and observe that event-based segments yield higher within-unit coherence and greater between-unit separability, leading to improvements in downstream tasks such as streaming QA and retrieval.
% Our finding: Event segmentation is not only intuitive, but empirically superior to time-based chunking. Motion spikes precede semantic drift by approximately 2s, suggesting that motion can serve as an early boundary cue.

\begin{figure}[t]
    \centering
    % ---- 上半部分 (a) ----
    \includegraphics[width=0.95\linewidth]{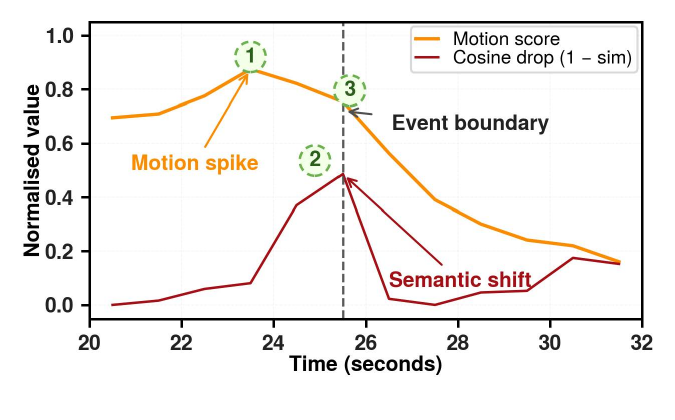}
    \vspace{3pt}
    
    % ---- 下半部分 (b) ----
    \begin{subfigure}[t]{0.32\linewidth}
        \centering
        \includegraphics[width=\linewidth]{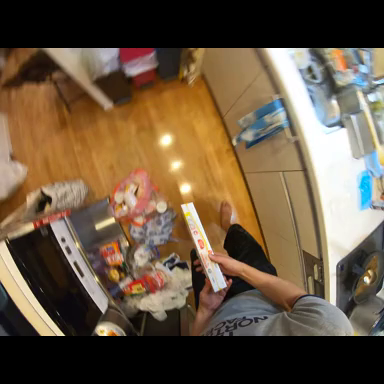}
        \caption{\text{23 s: Motion spike.}}
        \label{fig:motion_23s}
    \end{subfigure}
    \hfill
    \begin{subfigure}[t]{0.32\linewidth}
        \centering
        \includegraphics[width=\linewidth]{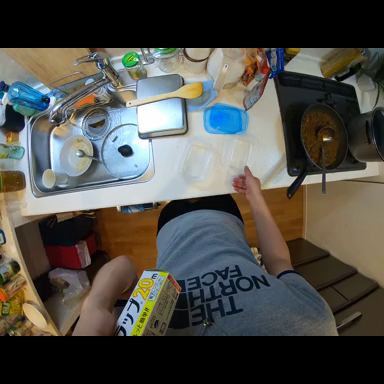}
        \caption{24.5 s: Transition phase.}
        \label{fig:motion_24_5s}
    \end{subfigure}
    \hfill
    \begin{subfigure}[t]{0.32\linewidth}
        \centering
        \includegraphics[width=\linewidth]{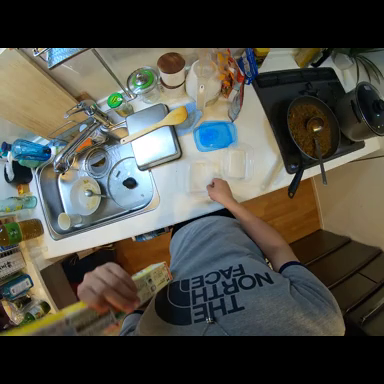}
        \caption{25 s: New action begins.}
        \label{fig:motion_25s}
    \end{subfigure}
    
    \vspace{-3pt}
    \caption{
        (\ul{\textbf{Top}}) Motion–semantic correlation curve. (\ul{\textbf{Bottom}}) Local motion–semantic transition: motion spikes (a) precede semantic drift (c) by $\sim$2s.
        Motion spikes precede semantic drift by approximately 2s, suggesting that motion can serve as an early cue for event boundaries.
    }
    \label{fig:fig3_combined}
    \vspace{-3pt}
\end{figure}

% \begin{figure}[t]
%     \centering
%     \includegraphics[width=1.0\linewidth]{figs/figure3b_new.pdf}
%     \caption{Motion spikes precede semantic drift by approximately 2s, suggesting that motion can serve as an early cue for event boundaries.\TODO{Leave space between index and time.}}
%     \label{fig:event_aware_vllm}
% \end{figure}

% Left: Motion score and cosine similarity change over a local video segment (23–31 s). A sharp motion spike appears earlier (~23 s), while the semantic shift (cosine drop) occurs later (~25–26 s), indicating that motion provides early boundary cues. Right: Corresponding frames illustrating the transition — (1) motion spike, (2) intermediate phase, and (3) emergence of a new action. This local view shows that motion spikes precede semantic drift by about 2 s.

\begin{figure}[t]
    \centering
    \includegraphics[width=\linewidth]{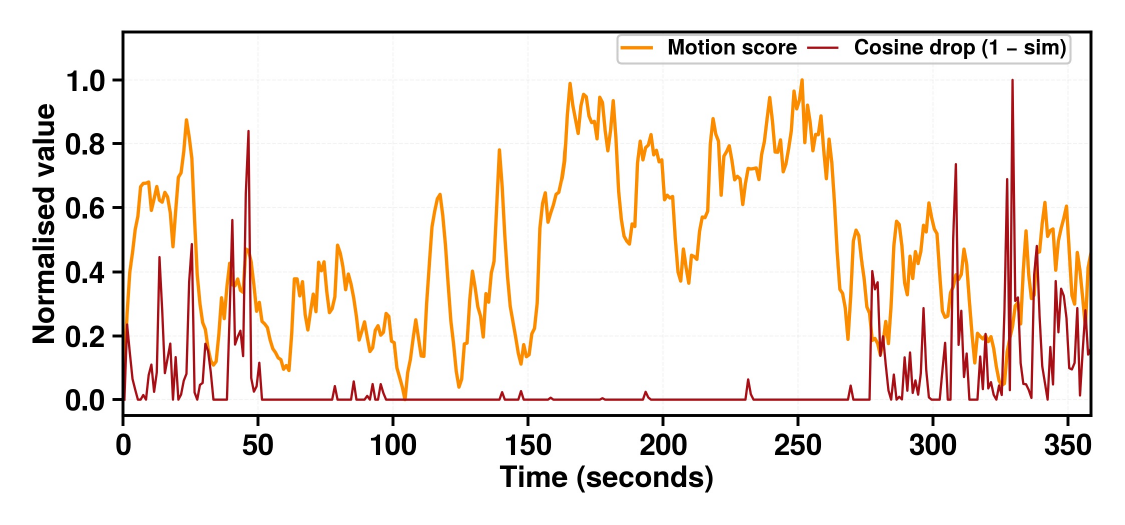}
    \caption{Frame-wise motion intensity vs. semantic similarity. Motion spikes often precede drops in cosine similarity, indicating that combining motion signals with semantic cues yields more accurate event boundaries.}
    \label{fig:motion_vs_similarity}
\end{figure}

\subsection{Cognitive Alignment: Our Boundary Model Matches Human Perception}
Event Segmentation Theory suggests that humans update mental event models when prediction error spikes\cite{basgol2024predictive,jung2025online}. We find a similar pattern in video streams: motion changes precede semantic drift (Figure~\ref{fig:fig3_combined}), acting as an early boundary cue, while semantic prediction error confirms the transition. Figure~\ref{fig:motion_vs_similarity} further plots frame-wise motion intensity against semantic similarity, showing that spikes in motion reliably precede drops in cosine similarity, which motivates combining motion and semantic cues for robust event boundary detection.Our finding: Natural boundaries arise when perceptual predictions fail—supporting an event-driven approach.

\section{Method}
We propose \textbf{Event-VStream}, a streaming video understanding framework that dynamically detects \text{semantic events}, maintains \text{Event Memory}, and performs language generation only when meaningful visual transitions occur, as summarized in Figure~\ref{fig:framework}
The system consists of three core components: 
\Circled{1} Event Boundary Detector that identifies state changes by integrating motion, semantic, and predictive cues; 
\Circled{2} Event Memory that consolidates event embeddings into a compact event-level memory bank for long-horizon streaming reasoning; and 
\Circled{3} Event-Driven Decoder that generates textual responses conditioned on the current event and the retrieved contextual memory. 
% The overall event-centric pipeline is illustrated in Figure~\ref{fig:process}.

\subsection{Streaming Pipeline}
Each incoming video stream $\mathbf{X} = \{x_1, x_2, \cdots, x_\texttt{T}\}$ is processed online. For each frame $x_t$, we extract a visual embedding $f_\texttt{t} = \mathbf{Enc}_\texttt{v}(x_\texttt{t})$ and maintain a running representation $\bar{f}$ for the current event. 

The \text{Event Boundary Detector} estimates a boundary probability $p_\texttt{t} = \mathbf{\sigma}(\mathbf{E}_\texttt{t})$ and emits an event token when $p_\texttt{t} > \tau_\texttt{t}$. 
Once triggered, the \textit{event memory} updates its stored embeddings $\mathcal{M} = \{E_1, \dots, E_k\}$, while the \textit{Event-Driven Decoder} consumes $\{E_k, \mathcal{M}\}$ to generate textual responses. 

The entire process operates in a single forward loop without access to future frames, ensuring strictly causal and real-time inference.

\begin{algorithm}[t]
\caption{\textbf{Event-VStream}: Event-centric streaming inference.}
\label{alg:eventvstream}
\textbf{Goal:} Detect semantic event boundaries from a video stream, update memory, and generate text only at transitions.\\
\textbf{Input:} Frames $\{x_t\}_{t=1}^T$, encoder $\mathrm{Enc}_v$,
language model $\mathrm{LM}$,
weights $w_{\text{sem}}, w_{\text{mot}}, w_{\text{pred}}$,
EMA rate $\rho$, threshold $\tau$.\\
\textbf{Output:} Event-level responses $\{y_k\}$.
\vspace{3pt}
\begin{algorithmic}[1]
\State Initialize memory $\mathcal{M} \leftarrow \varnothing$, event index $k \leftarrow 1$, start time $t_s \leftarrow 1$, running feature $\bar f \leftarrow \mathrm{Enc}_v(x_1)$
\For{$t = 1$ \textbf{to} $T$}
    \State $f_t \leftarrow \mathrm{Enc}_v(x_t)$
    \State Compute cue scores $s_t, \tilde m_t, c_t$ % e.g., using Eq.~(2)--(4)
    \State \textbf{Compute boundary score:}
    \[
        E_t \leftarrow
        w_{\text{sem}} (1 - s_t)
        + w_{\text{mot}} \,\tilde m_t
        + w_{\text{pred}} \, c_t
    \]
    \If{$E_t > \tau$}  \Comment{event boundary detected}
        \State $E_k \leftarrow \mathrm{Mean}(\{f_i\}_{i = t_s}^{t})$
        \State Update memory $\mathcal{M}$ and output $y_k \leftarrow \mathrm{LM}(E_k,\mathcal{M})$
        \State $k \leftarrow k + 1$
        \State Reset $\bar f \leftarrow f_t$, $t_s \leftarrow t + 1$
    \Else
        \State $\bar f \leftarrow (1 - \rho)\,\bar f + \rho f_t$
    \EndIf
\EndFor
\end{algorithmic}
\end{algorithm}

\subsection{Event Boundary and Representation Learning}
We convert continuous video streams into \emph{event-level} representations, which act as the core semantic units for streaming reasoning. An \textbf{event} is a temporally coherent segment in which visual semantics remain stable.

\paragraph{Preliminaries}
For each incoming frame $x_t$, we extract a visual embedding $f_t=\mathbf{Enc}_v(x_t)$.
All embeddings are $\ell_2$-normalized.
We maintain a running event representation $\bar f$ updated by an exponential moving average (EMA):
$\bar f \leftarrow (1-\rho)\,\bar f + \rho\, f_t$ with $\rho\in(0,1)$.
We denote the framewise motion magnitude by $m_t$ (e.g., mean optical-flow norm or frame-diff energy); unless otherwise specified, $m_t$ is min–max normalized within a sliding window.

\textbf{Event boundary score.} 
To detect event transitions, we integrate three complementary cues into an 
\emph{event boundary score}:
\begin{equation}
E_t = w_{\text{sem}} (1 - s_t)
    + w_{\text{mot}} \tilde{m}_t
    + w_{\text{pred}} c_t, 
\quad
p_t = \sigma(E_t),
\label{eq:event_score}
\end{equation}
where 
\(s_t = \cos(f_t, \bar{f})\),
\(\tilde{m}_t = \text{Norm}(m_t)\),
and 
\(c_t = \text{Norm}(\hat{c}_t)\).
The coefficients \(w_{\text{sem}}, w_{\text{mot}}, w_{\text{pred}} \ge 0\) 
are scalar hyperparameters that balance semantic drift, motion, and prediction error.

We mark frame \(t\) as an event boundary when its boundary probability exceeds
an adaptive threshold:
\begin{equation}
b_t = \mathbf{1}\!\left[p_t > \tau_t\right],
\label{eq:boundary-indicator}
\end{equation}
where $\tau_t$ is the adaptive threshold defined in Eq.~(\ref{eq:adaptive-threshold}).

\textbf{Causality of the prediction error.}
To avoid peeking at future frames, we define the one-step
prediction error in a causal form:
\begin{equation}
\hat{c}_t = \big\| g_\theta(f_{t-1}) - f_t \big\|_2^2.
\end{equation}
Here $g_\theta$ is instantiated as a lightweight three-layer
MLP that takes the previous frame embedding $f_{t-1}$ as
input and predicts the current embedding $f_t$. The predictor
is optimized with a self-supervised next-embedding $\ell_2$
loss on training videos, and then frozen for all experiments.
At inference time, we normalize the prediction error to obtain
the cue $c_t = \mathrm{Norm}(\hat{c}_t)$ used in
Eq.~\ref{eq:boundary-indicator}, enabling event-boundary
detection from raw video without manual annotations.
We then obtain a boundary probability $p_t = \sigma(E_t)$
and trigger a boundary when $p_t > \tau_t$.

\paragraph{Boundary cues}
The three terms in Eq.~\ref{eq:event_score} capture complementary aspects of event perception:
\begin{itemize}
    \item \textbf{Semantic drift} $(1-s_t)$ confirms a representation-level change and suppresses transient motion noise.
    \item \textbf{Motion cue} $\tilde m_t$ provides an early signal of abrupt physical transitions (camera pan/object motion).
    \item \textbf{Prediction error} $c_t$ reflects endogenous semantic shift when the next state becomes difficult to predict.
\end{itemize}
This design operationalizes the event-segmentation view that boundaries emerge when perceptual predictions fail.

\paragraph{Adaptive threshold}
In practice, $\tau_t$ can be fixed or adaptively modulated by short-term motion variance:
\begin{equation}
\tau_t \;=\; \tau_0\Big(1 + \eta \cdot \mathrm{Var}(m_{t-w:t})\Big),
\label{eq:adaptive-threshold}
\end{equation}
where $\eta$ controls temporal sensitivity and $w$ is a short history window.
This optional mechanism tightens/relaxes the boundary criterion under high/low dynamics; a fixed $\tau_t=\tau_0$ already works robustly.

\paragraph{Event construction}
When a boundary is triggered, we aggregate embeddings within the segment into a boundary-aware event token:
\begin{equation}
E_k \;=\; \frac{\sum_{i\in \text{seg}} w_i f_i}{\sum_{i\in \text{seg}} w_i},
\qquad 
w_i \propto \exp\!\Big(-\frac{|t_i-t_b|}{\sigma}\Big),
\label{eq:event_representation}
\end{equation}
where $t_b$ is the detected boundary time and $\sigma$ controls temporal sharpness.
This pooling preserves salient changes while suppressing redundancy, yielding compact tokens for downstream memory and reasoning.

\subsection{Event Memory}
The event memory $\mathcal{M} = \{E_1, E_2, \dots, E_k\}$ stores \emph{vision-side, event-level} embeddings for long-horizon reasoning. 
Unlike frame-wise buffering or time-uniform caches that accumulate raw frame features, our memory abstracts \emph{semantic events} as the basic unit, enabling compact storage and coherent retrieval.

\paragraph{Memory Update}
The memory bank is updated only when a new event token $E_k$ is formed. 
To prevent redundancy and drift, we adopt a lightweight \emph{merge-or-append} rule:
if the new event is highly similar to the most recent entry, we merge to stabilize the representation; otherwise, we append a new slot.
Formally, for the last entry $E_{\text{last}}$,
\begin{equation}
E_{\text{last}} \leftarrow
\begin{cases}
(1-\lambda)\,E_{\text{last}} + \lambda\,E_k, & \text{if } \cos(E_k, E_{\text{last}}) > \gamma_{\text{mem}},\\[4pt]
E_k, & \text{otherwise.}
\end{cases}
\label{eq:memory_update}
\end{equation}
Here, $\lambda \in (0, 1)$ controls the merge strength, and
$\gamma_{\text{mem}}$ is the redundancy threshold that decides
whether a new event should be merged into the last memory slot
or appended as a new one.

\paragraph{Discussion}
By operating at the event level rather than per-frame features, the event memory directly captures state transitions and avoids token-level duplication, enabling efficient and stable reasoning over extended streams.

\subsection{Event-Driven Streaming Decoding}
Rather than emitting text for every frame, decoding is triggered \emph{only at event boundaries}. 
Let $t_k$ be the $k$-th boundary time and $E_k$ the corresponding event token. 
We retrieve context $\mathcal{R}_k=\mathrm{Retrieve}(\mathcal{M},E_k)$ and generate an update 
\begin{equation}
    y_k \;=\; \mathrm{LM}(E_k,\mathcal{R}_k),
\end{equation}
While within boundaries, the model remains silent (the causal state is tracked without text generation).

To ensure stable pacing, we apply a simple hysteresis policy with minimum/maximum intervals $(\Delta_{\min},\Delta_{\max})$:
(i) boundaries within $\Delta_{\min}$ of $t_{k-1}$ are \emph{coalesced} into the current event (suppresses bursty updates), and 
(ii) if no boundary occurs for $\Delta_{\max}$, we emit a \emph{keep-alive} update using the latest state (prevents excessive silence).
This event-driven decoding eliminates repetitive descriptions of near-identical frames and yields coherent, context-aware updates aligned with human commentary rhythm.

\section{Experiment}
\subsection{Baselines and Experimental Setup}

\subsubsection{Datasets and Benchmarks}
We evaluate Event-VStream on online captioning and streaming video understanding in both mid-range and long-horizon settings. For open-world real-time reasoning, we adopt \textbf{OVOBench-Realtime}~\cite{li2025ovobench}, which measures semantic accuracy, responsiveness, and temporal coherence under continuous video input. OVOBench-Realtime provides a diverse suite of online tasks that stress a model's ability to operate in strictly causal conditions while maintaining context over long, untrimmed videos.

To assess long-horizon stability, we further construct a 2-hour egocentric evaluation suite based on Ego4D~\cite{grauman2022ego4d}. Among 9{,}821 Ego4D videos, only 112 exceed two hours ($\geq 7{,}200$~s). We select four unedited, continuous sequences (102--120 minutes each) covering daily activities such as cooking, indoor navigation, household interaction, shopping, and social 
communication. These streams exhibit irregular camera motion, spontaneous events, and highly variable pacing, and thus closely reflect the dynamics of real-world egocentric video.

Inspired by the evaluation protocol in
StreamingVLM\cite{xu2025streamingvlm}, our long-horizon evaluation relies on GPT-5 as an automatic judge, which may introduce biases in favor of certain linguistic styles; we partially mitigate this via bidirectional A/B testing and leave human studies for future work.

% \subsubsection{subsection{Baselines}
% We select strong baselines to compare with our \textbf{Event-Aware Video-Language Model (EAVLM)}. 
% For the streaming video understanding task, we include both proprietary and open-source models. 
% Specifically, we use \textbf{GPT-4o mini},to illustrate upper-bound commentary capability under powerful commercial systems. 
% For open-source counterparts, we compare against \textbf{StreamingVLM}~\cite{xu2025streamingvlm}, 
% \textbf{ReKV}~\cite{wang2024rekv}, \textbf{LiveCC-7B-Instruct}~\cite{chen2024videollmonline}, 
% and \textbf{Flash-VStream}~\cite{zhang2024Flash-VStream}. 
% Due to design constraints, GPT-4o mini and on \textbf{Inf-Streams-Eval} in the \textit{chunked} setting rather than the infinite-stream mode supported by our framework. 
% LiveCC-7B-Instruct and StreamingVLM are tested under both \textit{chunked} and \textit{infinite} configurations for fair comparison.

\subsubsection{Implementation details}
Our method can be applied on top of existing video–language models without retraining. We implement \textbf{Event-VStream} by augmenting the \textbf{VideoLLM-Online} framework~\cite{chen2024videollm} into a fully event-driven streaming pipeline. 
The implementation is model-agnostic and can interface with various vision-language backbones; here, we adopt \textbf{VideoLLM-Online} for a fair comparison. 
Instead of decoding at fixed frame intervals, the system continuously monitors motion and semantic drift to detect meaningful state transitions in real time. 
Each input video stream is sampled at 2\,FPS and encoded using the pretrained vision encoder from \textbf{VideoLLM-Online}. 
The extracted frame embeddings are dynamically grouped into semantically coherent events according to the boundary score $E_t$ (Eq.~\ref{eq:event_score}), which jointly integrates motion, semantic, and predictive cues.

We set the base similarity threshold to $\tau_0 = 0.96$
and the adaptive coefficient in Eq.~\ref{eq:adaptive-threshold} to $\eta = 0.03$,
following~\cite{chen2024videollmonline}.
To enhance robustness, we introduce \textbf{event-adaptive modulation}, which relaxes thresholds in stable scenes and tightens them under rapid motion, aligning event segmentation with real-world temporal dynamics. 
Each detected event embedding is $\ell_2$-normalized and stored in an event-level memory module for persistent retrieval and long-horizon reasoning.

All components, including event detection, memory update, and event-driven decoding, operate within a single forward loop without access to future frames, ensuring strictly causal, real-time inference. 
The system supports both finite-length and infinite-streaming modes. It achieves a stable throughput of approximately 17\,FPS on a single RTX~6000~Ada GPU, demonstrating efficient and scalable online processing under continuous video input.

\subsection{Streaming video understanding}
We evaluate \textbf{Event-VStream} across both short-term and long-horizon streaming settings, covering real-time QA, captioning, and long-duration stability.

\subsubsection{Long-Video Stability on Ego4D}
\begin{figure}[t]
    \centering
    \vspace{-0.6em}
    \includegraphics[width=0.95\linewidth]{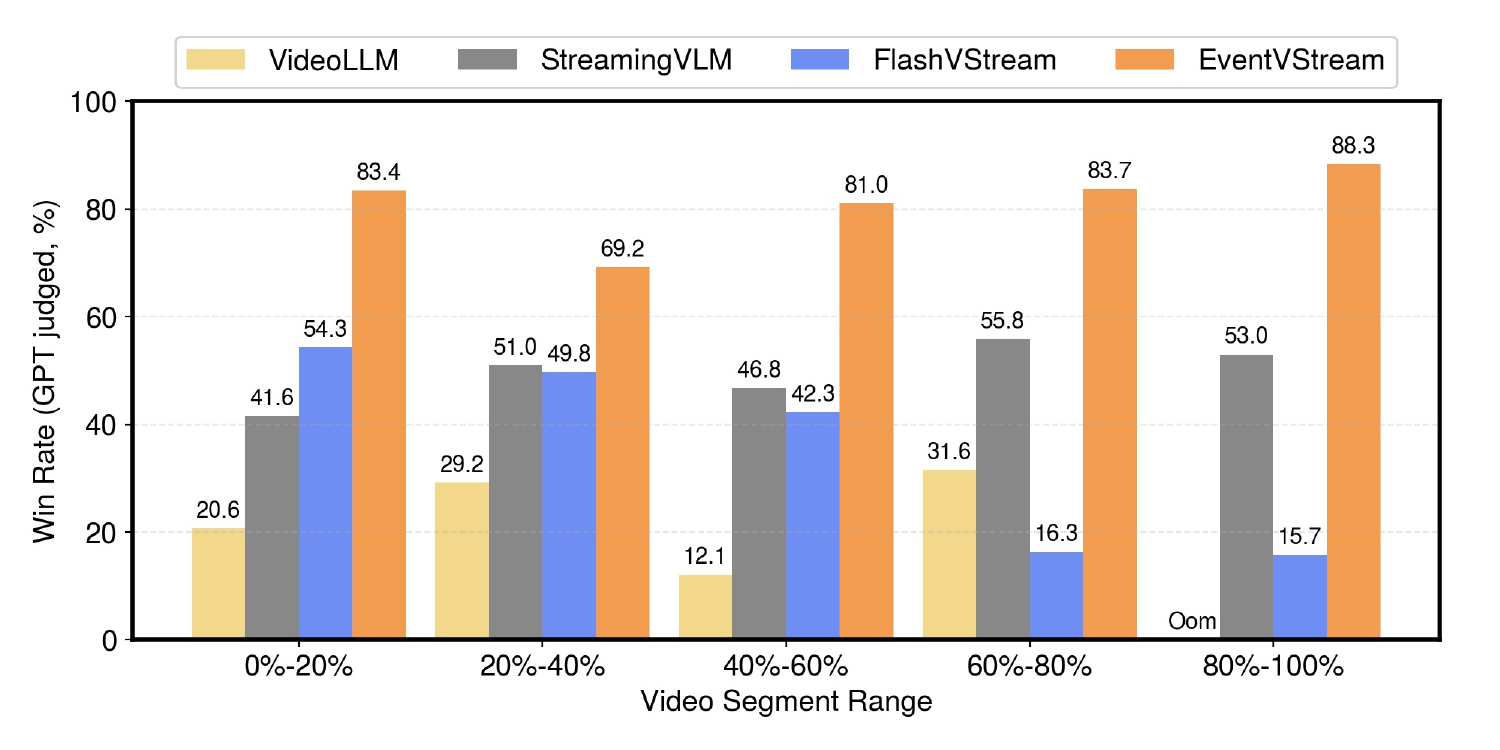}
    \vspace{-0.8em}
    \caption{
        \textbf{GPT5-based evaluation of long-stream stability.}
        Event-VStream maintains over 70\% GPT-5 win rate over 2-hour streams, demonstrating long-horizon stability.
    }
    \label{fig:stability_over_time}
    \vspace{-0.3em}
\end{figure}

To evaluate stability over extended durations, each 2-hour test video is divided into five 20\% segments, and GPT-5 win rates are computed within each segment(Figure~\ref{fig:stability_over_time}). 
As shown, StreamingVLM (gray) quickly degrades due to frequent cache refresh and loss of temporal continuity, while Flash-VStream (blue) shows unstable performance caused by redundant re-encoding of similar frames.VideoLLM-online (yellow) lacks KV-cache management and quickly collapses into repetitive 
\textbf{highly repetitive, low-quality outputs} before eventually running Oom.
In contrast, \textbf{Event-VStream (orange)} maintains around 70\% (up to 88.3\% in the final segment) win rate throughout all segments and remains stable even after two hours of continuous input. 
This demonstrates that event-level updates and memory consolidation effectively prevent context drift and forgetting, enabling coherent reasoning across unbounded video streams.
Qualitative inspection further shows that motion-only spikes trigger false updates in baselines, whereas our joint motion–semantic boundary detector suppresses them, preserving narrative consistency.

\subsubsection{OVOBench-Realtime Accuracy}

We further evaluate \textbf{Event-VStream} on \textbf{OVOBench-Realtime}~\cite{li2025ovobench}, a benchmark targeting open-world real-time video reasoning. As shown in Table~\ref{tab:ovo_realtime}, Event-VStream achieves an average score of 28.15, yielding a +10.4 point absolute gain over its  VideoLLM-Online-8B baseline (17.73) under the same 2 FPS input and visual encoder. This suggests that the improvement primarily comes from the proposed event-driven streaming mechanism rather than increased model capacity.

Although Event-VStream is instantiated on the VideoLLM-Online baseline~\cite{chen2024videollmonline} built on a general-purpose LLaMA-3-8B\cite{meta2024llama3} text-only language model, its performance is only 0.83 points lower than \textbf{Flash-VStream-7B} (28.37\%), which is built on a vision-specialized LLaVA-based architecture~\cite{liu2023visualinstructiontuning}. This narrow gap indicates that the \textbf{event-centric streaming design}, rather than a specialized backbone, accounts for most of the gains.

% \subsubsection{Question-answering performance}
% summarize Event-VStream’s question-answering accuracy across the three evaluation categories in StreamingBench—Real-Time Visual Understanding, Omni-Source Understanding, and Contextual Understanding—covering 18 subtasks in total. Event-VStream substantially surpasses prior state-of-the-art online methods, achieving average gains of 1.8 , 7.7 , and 2.3  across the three categories, respectively. Compared with its base model LLaVA-OneVision-7B, Event-VStream achieves consistent accuracy improvements while operating on only 30 of visual tokens through event-driven segmentation and memory recycling. In contrast, ReKV [31] shows a uniform drop in performance across all tasks. These results highlight the advantage of Event-VStream’s event-centric processing pipeline for streaming video question-answering, which balances semantic retention and efficiency by reasoning at the event level rather than the frame level.

% \begin{table*}[t]
% \centering
% \caption{\textbf{Open-source Online Video LLMs.} 
% Evaluation results on eight streaming understanding tasks grouped into 
% Omni-Source and Contextual categories.}
% \vspace{-3pt}
% \resizebox{\linewidth}{!}{
% \begin{tabular}{lccccccccccc}
% \toprule
% \textbf{Model} & \textbf{Frames} &
% \multicolumn{5}{c}{\textbf{Omni-Source Understanding}} &
% \multicolumn{5}{c}{\textbf{Contextual Understanding}} \\
% \cmidrule(lr){3-7} \cmidrule(lr){8-12}
% & & ER & SCU & SD & MA & All & ACU & MCU & SQA & PO & All \\
% \midrule
% Flash-VStream-7B [CVPR 24] & -- & 25.91 & 24.90 & 25.60 & 28.40 & 26.00 & 24.80 & 25.20 & 26.80 & 1.96 & 24.12 \\
% VideoLLM-online-8B [CVPR 24] & 2 fps & 31.20 & 26.51 & 24.10 & 32.00 & 28.45 & 24.19 & 29.20 & 30.80 & 3.92 & 26.55 \\
% StreamingVLM [CVPR 25] & 1 fps & 35.46 & 25.26 & 38.57 & 43.34 & 35.66 & 39.62 & 27.65 & 34.80 & 25.34 & 33.61 \\
% ReKV-7B [ICLR 25] & 0.5 fps & 38.80 & 24.80 & 39.60 & 46.40 & 37.40 & 31.20 & 30.40 & 30.40 & 30.80 & 30.70 \\
% \rowcolor{yellow!20}
% \textbf{Event-VStream (Ours)} & 0.5 fps & \textbf{44.40} & -- & \textbf{30.40} & \textbf{48.80} & \textbf{60.80} & \textbf{46.10 \small{(7.70~$\uparrow$)}} & 38.80 & \textbf{36.80} & 32.00 & \textbf{35.00 \small{(2.26~$\uparrow$)}} \\
% \bottomrule
% \end{tabular}
% }
% \vspace{-5pt}
% \label{tab:open_source_llm}
% \end{table*}

\begin{table*}[t]
\centering
\small
\setlength{\tabcolsep}{6pt}   %
\renewcommand{\arraystretch}{1.12}

\caption{\textbf{Open-source Online Video LLMs.}
Evaluation results on streaming understanding tasks (OVO-Realtime).}
\label{tab:ovo_realtime}

\begin{tabularx}{\linewidth}{l c *{7}{>{\centering\arraybackslash}X}}
\toprule
Model & Frames & OCR & ACR & ATR & STU & FPD & OJR & Avg. \\
\midrule
Flash-VStream-7B(CVPR24) & 1 fps &
24.16 & 29.36 & 28.45 & 33.71 & 25.74 & 28.80 & 28.37 \\
VideoLLM-online-8B(CVPR24) & 2 fps &
8.05 & 23.85 & 12.07 & 14.04 & 45.54 & 18.80 & 17.73 \\
\rowcolor{yellow!20}
\textbf{Event-VStream (Ours)} & 2 fps &
24.38 & 29.36 & 27.59 & 25.09 & 33.66 & 28.80 & 28.15 \\
\bottomrule
\end{tabularx}
\end{table*}

% ============================================
% Table: Overall accuracy on OVOBench (Realtime)
% ============================================

% \begin{table}[t]
% \centering
% \small
% \setlength{\tabcolsep}{5pt}
% \renewcommand{\arraystretch}{1.1}
% \caption{
% \textbf{Overall accuracy on OVOBench (Realtime).}
% Our \textbf{Event-VStream} consistently surpasses existing online video-language models,
% achieving a new state-of-the-art total accuracy of \textbf{61.96\%}. The substantial gain over prior streaming frameworks highlights the strength of ourEvent-VStream design in maintaining temporal coherence under long-horizon inputs.
% % \TODO{Caption of table should be on the top.}
% }
% \begin{tabular}{lcc}
% \toprule
% \textbf{Model} & \textbf{OVOBench (Realtime)} & \textbf{Gain} \\
% \midrule
% VideoLLM & 54.23 & — \\
% StreamVLM & 55.74 & — \\
% Flash-VStream & 57.38 & — \\
% \rowcolor{yellow!20}
% \textbf{Ours (Event-VStream)} & \textbf{61.96} & \textbf{4.58~$\uparrow$} \\
% \bottomrule
% \end{tabular}

% \label{tab:ovobench_main}
% \end{table}

% ============================================
% Table: Overall accuracy on OVOBench (Realtime)
% ============================================

\begin{figure}[t]

    \centering
    \includegraphics[width=\linewidth]{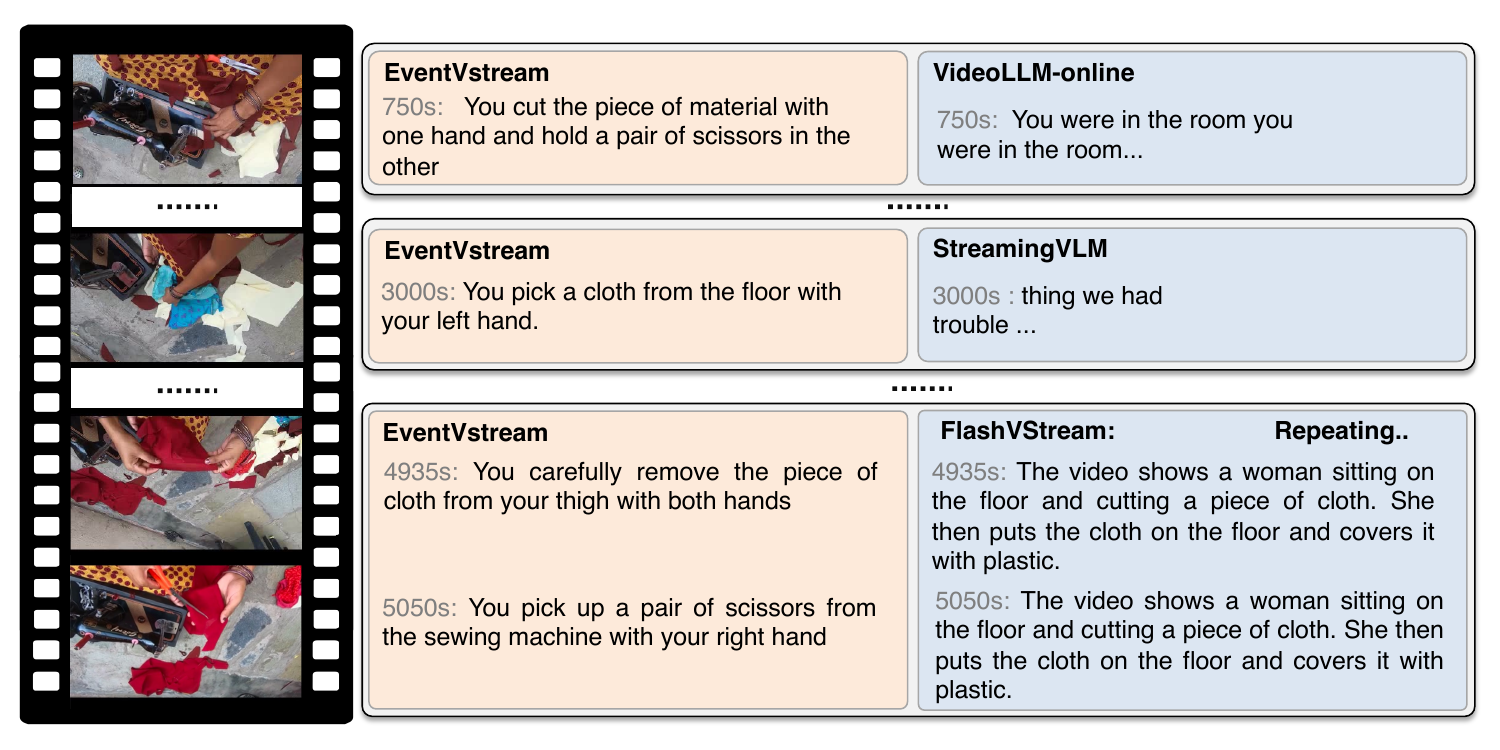}
    \caption{
        \textbf{Existing streaming VLMs fail for different reasons:} 
         StreamingVLM refreshes at fixed intervals and breaks temporal continuity, Flash-VStream clusters too coarsely and accumulates redundant tokens that pollute the KV cache, and VideoLLM lacks any cache management and quickly collapses into repetition or runs Oom. In contrast, Event-VStream performs updates only at true semantic events, keeps the cache compact by selecting representative frames, and maintains a coherent, bounded event-level memory over multi-hour streams.}
    \label{fig:generation_res}
\end{figure}
\vspace{-3mm}

\subsection{Qualitative Analysis}
Figure~\ref{fig:generation_res} compares long-horizon captioning behavior across streaming VLMs. \textbf{VideoLLM-Online}, without any cache management, quickly degenerates into repetitive loops as its KV cache grows. \textbf{StreamingVLM} refreshes at fixed intervals, producing short, fragmented sentences and breaking temporal continuity. \textbf{Flash-VStream} repeatedly re-encodes nearly identical frames, yielding long redundant descriptions that gradually pollute the KV cache. In contrast, \textbf{Event-VStream} triggers decoding only at jointly detected motion–semantic event boundaries, producing compact, well-formed updates that align with true state changes and preserve coherent narration over multi-hour streams. These qualitative results highlight event-driven decoding as a simple but effective remedy for repetition, fragmentation, and redundancy in streaming video captioning.

\subsubsection{Efficiency Tests}
\begin{figure}[t]
    \centering
    \includegraphics[width=\linewidth]{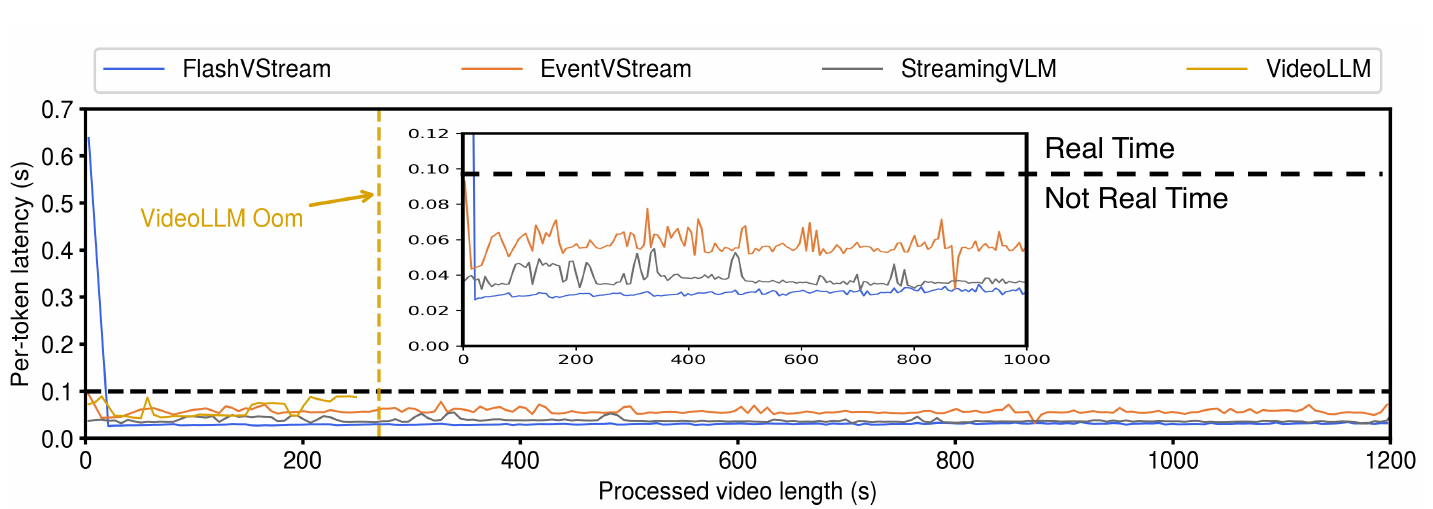}
    \caption{
        \textbf{Per-token generation latency comparison.} 
       Event-VStream maintains sub-0.1s latency for most frames, while StreamingVLM stays steady but higher; aligned traces and box plots highlight Event-VStream’s lower average latency with occasional spikes versus StreamingVLM’s more uniform delay.}
    \label{fig:generation_latency}
\end{figure}

We further compare the computational efficiency of \textbf{Flash-VStream}, \textbf{StreamingVLM}, \textbf{VideoLLM-Online}, and \textbf{Event-VStream} by measuring per-token generation latency on the 2-hour Ego4D streams. (Figure~\ref{fig:generation_latency})plots per-token generation latency as a function of processed video length, with a dashed line at 0.1 s/token indicating the real-time threshold. For each model, we record the time required to generate every output token and plot latency as a function of processed video length, with a dashed line at 0.1\,s/token indicating the real-time threshold. After an initial warm-up, Flash-VStream attains the lowest steady-state latency ($\approx$~0.03–0.04\,s/token), and StreamingVLM remains slightly higher yet still comfortably within the real-time regime. VideoLLM-Online operates close to the 0.1\,s/token boundary but its latency gradually increases and the model runs out of memory after roughly 300\,s, exposing the cost of uncompressed frame-wise processing. \textbf{Event-VStream} incurs a modest overhead from event-boundary detection and memory retrieval, yet maintains stable sub-0.1\,s latency across the full 2-hour streams, with most tokens in the 0.05–0.08\,s range and no observable drift over time. This indicates that event-driven updates and compact event-level memory preserve real-time performance over long horizons while avoiding the slowdown and eventual failure characteristic of fully frame-centric designs.

% In 6_ablation.tex
\section{Ablation}
To understand the contribution of each cue in the event-boundary estimation, we conduct a detailed ablation over motion, semantic, and prediction components in Table~\ref{tab:boundary_ablation}.Removing any cue degrades both segmentation quality and downstream caption performance, confirming their complementarity. Without motion, the model becomes less sensitive to rapid physical transitions and delays boundary detection. Dropping the semantic term causes the largest drop in GPT-5 caption win rate, as transient motions trigger false updates and fragment the narrative. Removing prediction error weakens the model's anticipatory ability, leading to higher latency and larger segmentation error. The full model, which combines all three cues, achieves the best trade-off between caption quality and efficiency.

\vspace{-0.8em}
\begin{table}[!t]
\centering
\setlength{\tabcolsep}{5pt}
\renewcommand{\arraystretch}{1.15}

\caption{
\textbf{Ablation on boundary cues (motion, semantic, prediction).}
Removing any cue reduces caption quality and increases latency,
while the full model achieves the best trade-off between quality and efficiency.
}
\label{tab:boundary_ablation}

\vspace{0.3em}

% \resizebox{\linewidth}{!}{
\begin{tabular}{p{2cm}>{\centering}p{0.4cm}>{\centering}p{0.4cm}>{\centering}p{0.4cm}p{1.6cm}p{1.5cm}}
\toprule
\text{Variant} & Mot. & Sem. & Pred. &
Cap.Win(\%)$\uparrow$ & Latency(s)$\downarrow$ \\
\midrule
w/o Motion     & \xmark & \cmark & \cmark & \centering 11.8 & 0.190 \\
w/o Semantic   & \cmark & \xmark & \cmark & \centering 37.5 & 0.173 \\
w/o Prediction & \cmark & \cmark & \xmark & \centering 46.7 & 0.187 \\
\rowcolor{yellow!20}
\textbf{Full (Ours)} & \cmark & \cmark & \cmark &
\centering \textbf{68.1} & \textbf{0.093} \\
\bottomrule
\end{tabular}
% }
\end{table}

\FloatBarrier

\section{Conclusion}
Event-VStream demonstrates that \textbf{representing streaming video as a sequence of discrete events} is an effective way to address redundancy and forgetting in online video understanding. By combining (i) event-based selective updates that trigger decoding only at meaningful state transitions, (ii) a lightweight event-level memory that merges redundant events into compact, persistent representations, and (iii) an event-driven decoding policy with simple pacing control, the framework achieves coherent long-horizon reasoning while maintaining real-time latency across multi-hour streams. On OVOBench-Realtime and 2-hour Ego4D evaluations, Event-VStream yields substantial gains over a VideoLLM-Online-8B baseline and approaches the performance of Flash-VStream-7B despite using a general-purpose LLaMA-3-8B text model.

Future work includes incorporating audio and speech signals to support multimodal event detection and extending the memory mechanism to multi-scale temporal reasoning in more complex real-world streams.

\newpage
{
    \small
    \bibliographystyle{ieeenat_fullname}
    \bibliography{main}
}


% \newpage
% {
%     \small
%     \bibliographystyle{ieeenat_fullname}
%     \bibliography{main}
% }


\end{document}